\documentclass[10pt, conference, compsocconf]{IEEEtran}
\usepackage[noadjust]{cite}
\usepackage{hyperref}

\ifCLASSINFOpdf
\else
\fi
\usepackage[cmex10]{amsmath}

\usepackage[utf8]{inputenc}
\usepackage[T1]{fontenc}
\usepackage{textcomp}

\usepackage{amsmath,array,graphicx}

\usepackage[caption=false,font=footnotesize]{subfig}
\usepackage{fixltx2e}

\usepackage{float}

\begin{document}
%
\title{Treepedia 2.0: Applying Deep Learning for\\ Large-scale Quantification of Urban Tree Cover}


\author{	
\IEEEauthorblockN{
Bill Yang Cai\IEEEauthorrefmark{1},
Xiaojiang Li\IEEEauthorrefmark{2}, 
Ian Seiferling\IEEEauthorrefmark{3},
and Carlo Ratti\IEEEauthorrefmark{4}}
\IEEEauthorblockA{Senseable City Lab\\
Massachusetts Institute of Technology\\
Cambridge, MA, USA\\
Email: \IEEEauthorrefmark{1}billcai@mit.edu,
\IEEEauthorrefmark{2}xiaojian@mit.edu,
\IEEEauthorrefmark{3}ianseifs@mit.edu,
\IEEEauthorrefmark{4}ratti@mit.edu}
}


%


\maketitle

\begin{abstract}
Recent advances in deep learning have made it possible to quantify urban metrics at fine resolution, and over large extents using street-level images. Here, we focus on measuring urban tree cover using Google Street View (GSV) images. First, we provide a small-scale labelled validation dataset and propose standard metrics to compare the performance of automated estimations of street tree cover using GSV. We apply state-of-the-art deep learning models, and compare their performance to a previously established benchmark of an unsupervised method. Our training procedure for deep learning models is novel; we utilize the abundance of openly available and similarly labelled street-level image datasets to pre-train our model. We then perform additional training on a small training dataset consisting of GSV images.  We find that deep learning models significantly outperform the unsupervised benchmark method. Our semantic segmentation model increased mean intersection-over-union (IoU) from 44.10\% to 60.42\% relative to the unsupervised method and our end-to-end model decreased Mean Absolute Error from 10.04\% to 4.67\%. We also employ a recently developed method called gradient-weighted class activation map (Grad-CAM) to interpret the features learned by the end-to-end model. This technique confirms that the end-to-end model has accurately learned to identify tree cover area as key features for predicting percentage tree cover. Our paper provides an example of applying advanced deep learning techniques on a large-scale, geo-tagged and image-based dataset to efficiently estimate important urban metrics. The results demonstrate that deep learning models are highly accurate, can be interpretable, and can also be efficient in terms of data-labelling effort and computational resources.

\end{abstract}


%
\IEEEpeerreviewmaketitle

\section{Introduction}
Recent advances in deep learning, particularly in deep convolutional neural networks (DCNN), provide researchers with a powerful tool to learn complex representations of unstructured data such as images. Since DCNNs beat previous benchmarks from shallow networks in competitions such as ImageNet, computer vision benchmark competitions have been dominated by researchers using deep learning methods \cite{GUO201627}. Along with the development of deep learning, the provision of large-scale datasets has greatly catalyzed deep learning research by providing the data they require to learn complex data representations, and establishing benchmark test datasets and standard evaluation metrics to compare performance across different methods \cite{russakovsky2015imagenet}.

\subsection{Traditional approaches to quantify urban tree cover}
Urban greenery, and in particular trees, attract substantial interest in both academic research and urban planning since they provide wide-ranging services in cities such as carbon sequestration and oxygen production \cite{Nowak2007220}, and heat island effect mitigation \cite{lafortezza2009benefits}. Furthermore, the perception of urban greenery has a significant influence on the visual appeal of streets \cite{bain2012living}, such that urban greenery programs have generally received support from local residents \cite{jim2006perception}.
\begin{figure}[h]
	\centering
	\resizebox{\columnwidth}{!}{%
		\includegraphics[scale=0.5]{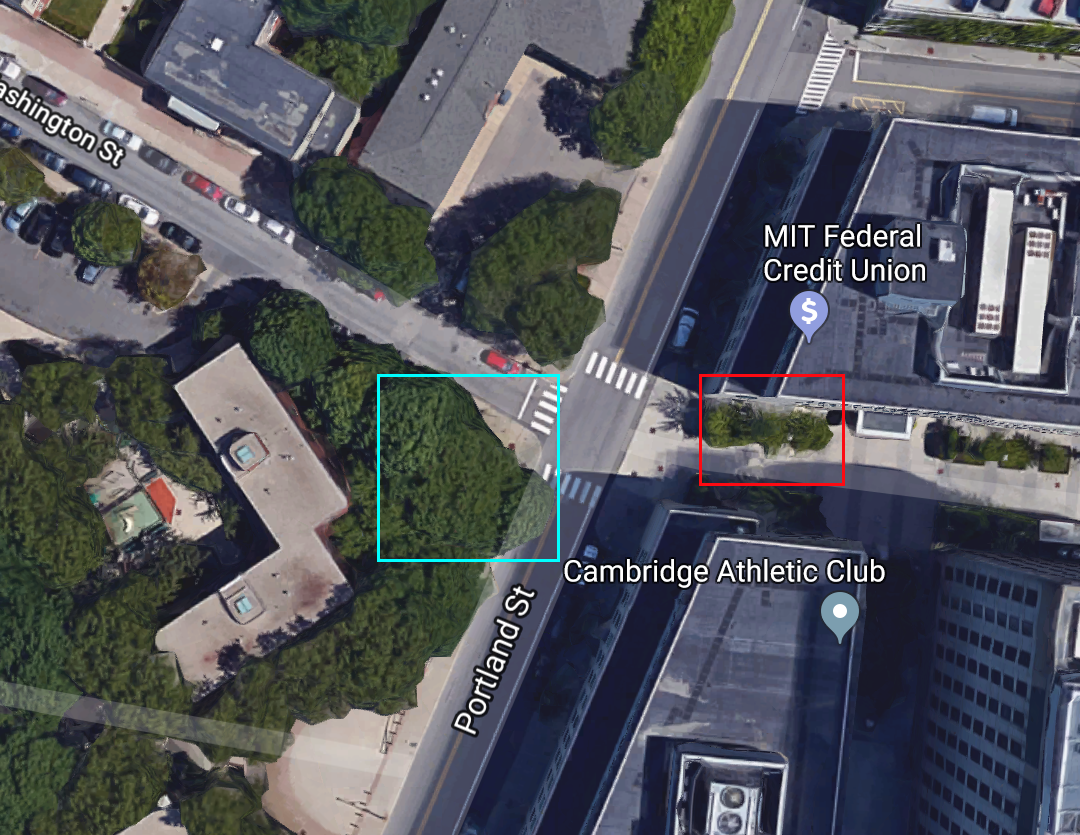}}
	\caption{Satellite image of junction between Portland and Washington St in Cambridge, MA. Trees highlighted in red have less dense canopies than tress highlighted in blue.}
\end{figure}\\
Traditional approaches to map vegetation cover mapping have relied on either overhead view remote sensing images or photographs, and surveyed data obtained via on-ground fieldwork. Yang et al \cite{yang2009can} argues that methods based on overhead view images are limited in their ability to represent the profile view of street vegetation that people experience on the streets. We provide a comparison between overhead and street view images (Fig. 1 and Fig. 2 respectively). The overhead view image does not show substantial differentiation between the spread and density of tree canopy along the pedestrian walkway on Washington St (red box in Fig. 1) and the spread and density of tree canopy at the opposite end of the road junction (blue box in Fig. 1). The street level images clearly portray differences between the tree canopy density of these two groups of trees (red and blue boxes, respectively, in Fig. 2). On the other hand, fieldwork images are limited in their coverage and are costly to acquire at citywide scales.
\begin{figure}[h]
	\centering
	\resizebox{0.45\columnwidth}{!}{%
		\includegraphics[scale=0.5]{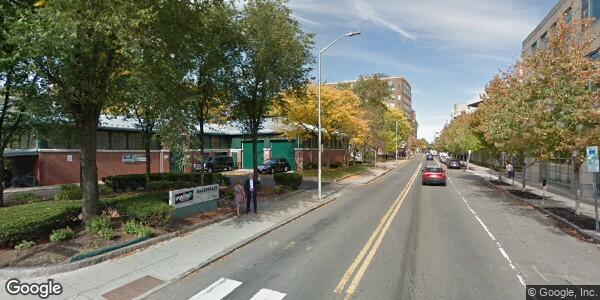}}
	\resizebox{0.45\columnwidth}{!}{%
		\includegraphics[scale=0.5]{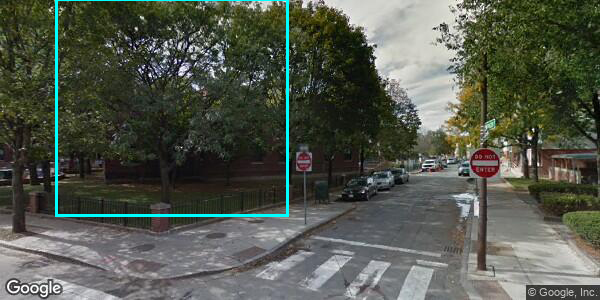}}\\
	\vspace{1mm}
	\resizebox{0.45\columnwidth}{!}{%
		\includegraphics[scale=0.5]{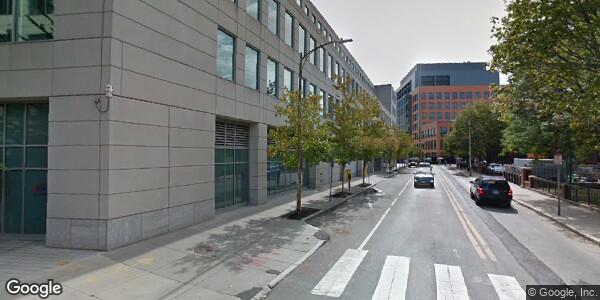}}
	\resizebox{0.45\columnwidth}{!}{%
		\includegraphics[scale=0.5]{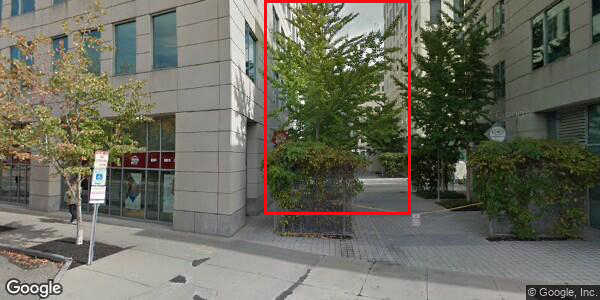}}
	\caption{Google Street View images of identical Portland and Washington St junction taken at different headings. Trees highlighted in red have less dense canopies than tress highlighted in blue.}
\end{figure}\\
The advent of large-scale and open-source street-level image datasets such as Google Street View (GSV) and Baidu Maps street view service provide researchers a new tool for assessing urban vegetation cover \cite{li2015assessing}. In addition to their extensive coverage across most of North America, Europe and East Asia excluding China, GSV images are collected and processed to obtain high-resolution, precisely standardized and located images of streetscapes \cite{anguelov2010google}. Furthermore, the recent explosion of research in autonomous vehicle technology has resulted in an abundance of large-scale labelled street-level image datasets such as the Cityscapes dataset \cite{cordts2016cityscapes} and Mapillary Vistas. Thus, GSV represents a comprehensive image dataset of the urban landscape that meets the training requirements of deep learning models.

To quantify urban tree cover, Yang et al \cite{yang2009can} and Li. et al \cite{li2015assessing} utilize the Green View Index (GVI) metric. The GVI metric measures street-level urban tree cover cover by averaging the quantity of identified tree canopy in street-level images taken at different headings and pitches from the same geographical location. To identify tree canopy cover in these images, Li. et al \cite{li2015assessing} uses an unsupervised mean shift image segmentation algorithm \cite{comaniciu2002mean} to measure the number of green pixels as a percentage of total pixels in a GSV image. While the benchmark unsupervised method does not require training data, segmentation results from this method are influenced by the presence of shadows and the illumination condition in the GSV images. Green features other than vegetation are also misidentified as urban tree canopies with this method. Seiferling et al. \cite{seiferling2017green} first applied supervised learning to quantify vegetation cover in GSV images, but their method involved pre-computing image features. To address these shortcomings, we make use of recent advances in supervised DCNN methods by training and testing two models. We train a DCNN semantic segmentation model to produce a densely segmented image. and we also train a DCNN end-to-end model that directly estimates GVI in GSV images.

To validate accuracy in estimating GVI, Li. et al \cite{li2015assessing} 
used 33 labelled GSV images and calculated the Pearson’s correlation coefficient between GVI from labelled images and model estimates. In this paper, we use a larger test set of 100 GSV images randomly selected across cities in our study area. We propose two additional metrics to validate prediction accuracy of our models: (1) Mean Intersection-over-Union (IoU) is a standard metric used in the computer vision field and captures the accuracy of the location of assigned "green vegetation" pixel labels. (2) Mean Absolute Error calculates the mean of absolute differences in labelled GVI and estimated GVI across images in the test GSV dataset. These two metrics provide a direct measure of accuracy in locating urban tree cover and measuring GVI respectively, while Pearson’s correlation coefficient only provides information on the strength of co-movement between labelled GVI and estimated GVI.

\subsection{Deep learning in quantifying urban canopy cover}
The use of DCNN in urban ecology and geographic information science (GIS) is in its early stages and relies mostly on overhead, often satellite, imagery. A recent survey by Cheng et al.'s of the use of DCNN in recent remote sensing literature \cite{cheng2017remote} concluded that datasets of non-overhead images, especially of geo-tagged and profile-perspective social media photos provide a larger-scale and more visually diverse collection of images for deep learning algorithms as compared to imagery collected from satellites and airborne surveys. 

Wegner et al \cite{wegner2016cataloging} is an example of recent work that employed DCNN models to street-level images in the urban ecology field. The authors developed a sophisticated method that uses DCNN models on street-level and overhead images to accurately locate and classify tree species. While the information obtained about urban trees from Wegner et al’s method is rich in terms of mapping individual trees \cite{wegner2016cataloging}, our DCNN method of estimating the GVI metric produces a much faster aggregate measure of tree cover without extensive training data. Therefore, our method is more easily applied at the city-scale with limited computational resources. Moreover, aggregate measure of tree cover along streets, in and of itself, provides rich and useful information to a variety of stakeholders.

Early applications of DCNN in urban ecology and geospatial sciences have mainly focused on the accuracy of DCNN models. To date, there have been little to no DCNN and computer vision applications in urban ecology or GIS that have sought to understand the features learned and identified by DCNN models. As an active and important area of research amongst computer vision scientists, tools to interpret and "peer into" black-box DCNN models are important to visually validate trained models, ascertain that trained models do not contain significant biases, and convince researchers from fields outside of computer vision about the decision process of black-box DCNN models. Therefore, for DCNN models that we employ, we use a visually interpretable method called gradient-weighted class activation map (Grad-CAM) \cite{selvaraju2016grad} to understand the learned features extracted by our trained DCNN models.
\section{Methodology}
\subsection{Obtaining GSV training-validation-test dataset}
We choose Cambridge (Massachusetts, USA), Johannesburg (South Africa), Oslo (Norway), S$\widetilde{\mbox{a}}$o Paulo (Brazil), and Singapore (Singapore) as cities included in our training and test sets. The 5 cities chosen to for this paper are located in differing climatic regions, have various tree species assemblages and forms, and have varying levels of green cover according to the GVI scores estimated by the benchmark \textit{Treepedia} unsupervised segmentation method. This diversity amongst cities should imply model generalization if strong predictive results are found. We use OpenStreetMap street network extracts of the 5 sample cities, and sample points 1 km apart on road networks. We then query the Google Maps API to determine the availability of GSV taken at sampled points. From each of the 5 cities, we randomly select 100 available GSV images each to form a training-validation-test set. We then divide the 500 image dataset into a 100 image test set, 320 image training set and a 80 image validation set. We produce manual labels by carefully tracing all vertical vegetation in the images for all 500 images (Fig. 3).

\begin{figure}[h]
	\centering
	\resizebox{0.45\columnwidth}{!}{%
		\includegraphics[scale=0.5]{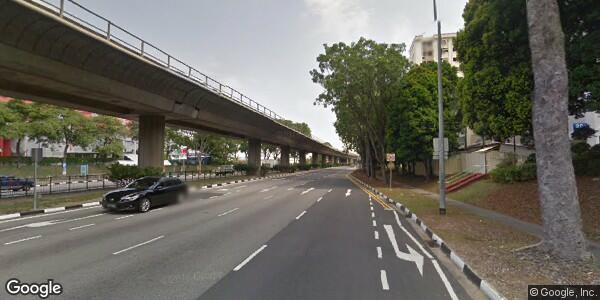}}
	\resizebox{0.45\columnwidth}{!}{%
		\includegraphics[scale=0.5]{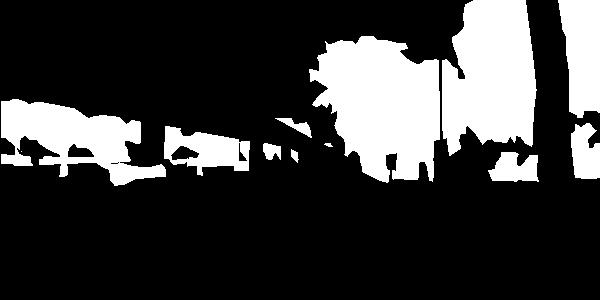}}\\
	\vspace{1mm}
	\resizebox{0.45\columnwidth}{!}{%
		\includegraphics[scale=0.5]{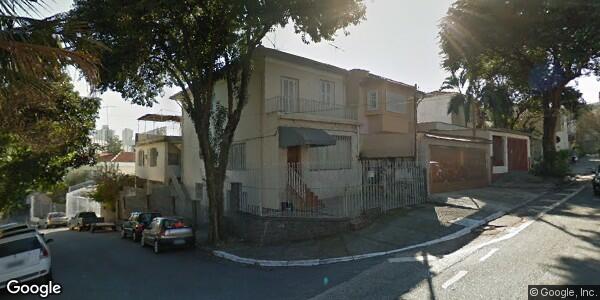}}
	\resizebox{0.45\columnwidth}{!}{%
		\includegraphics[scale=0.5]{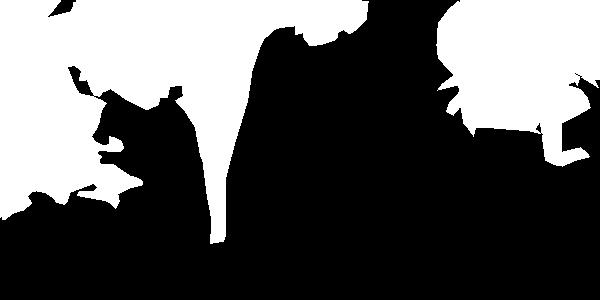}}
	\caption{Top: GSV image in Singapore and associated vegetation labels. Bottom: GSV image in  S$\widetilde{\mbox{a}}$o Paulo and associated vegetation labels.}
\end{figure}
\subsection{Augmenting training data}
We augment our model training by first using the high-quality, pixel-labelled Cityscapes dataset to initially train our DCNN model. The Cityscapes dataset is currently one of the largest, most extensive and most richly annotated image dataset of city streetscapes to date \cite{cordts2016cityscapes}. The finely labelled subset of the Cityscapes dataset contains over 5000 images taken from vehicle-mounted cameras placed in vehicles and sampled across numerous German cities. We also utilize the class labels for the Cityscapes dataset, which contain a class label for vertical vegetation. We convert the Cityscapes dataset by collapsing the original multi-class labels into binary labels for vegetation and non-vegetation. By first training our models on the larger Cityscapes dataset, we increase our training dataset with the aim of increasing our model performance. 
\begin{figure}[h]
	\centering
	\resizebox{0.45\columnwidth}{!}{%
		\includegraphics[scale=0.5]{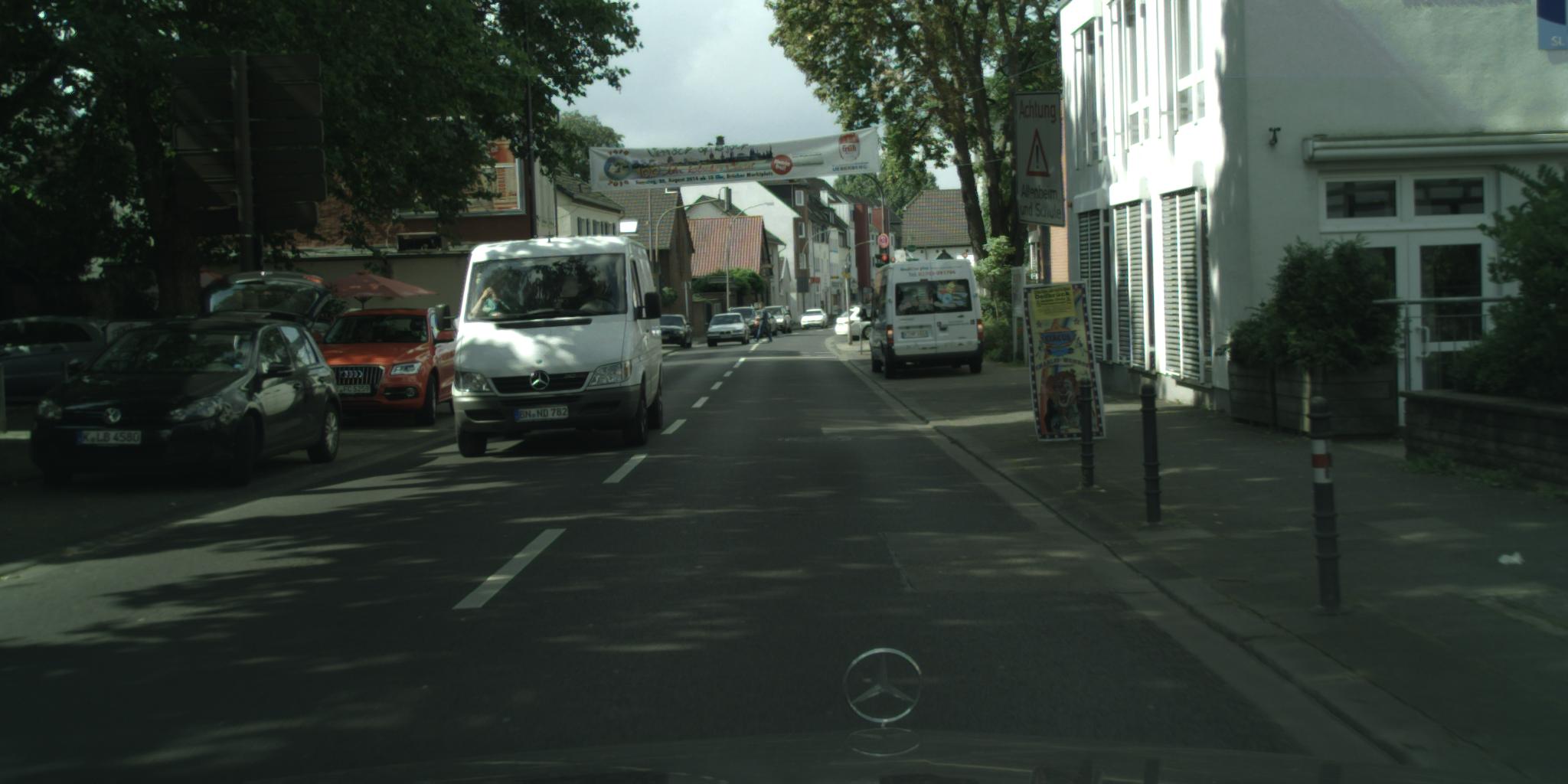}}
	\resizebox{0.45\columnwidth}{!}{%
		\includegraphics[scale=0.5]{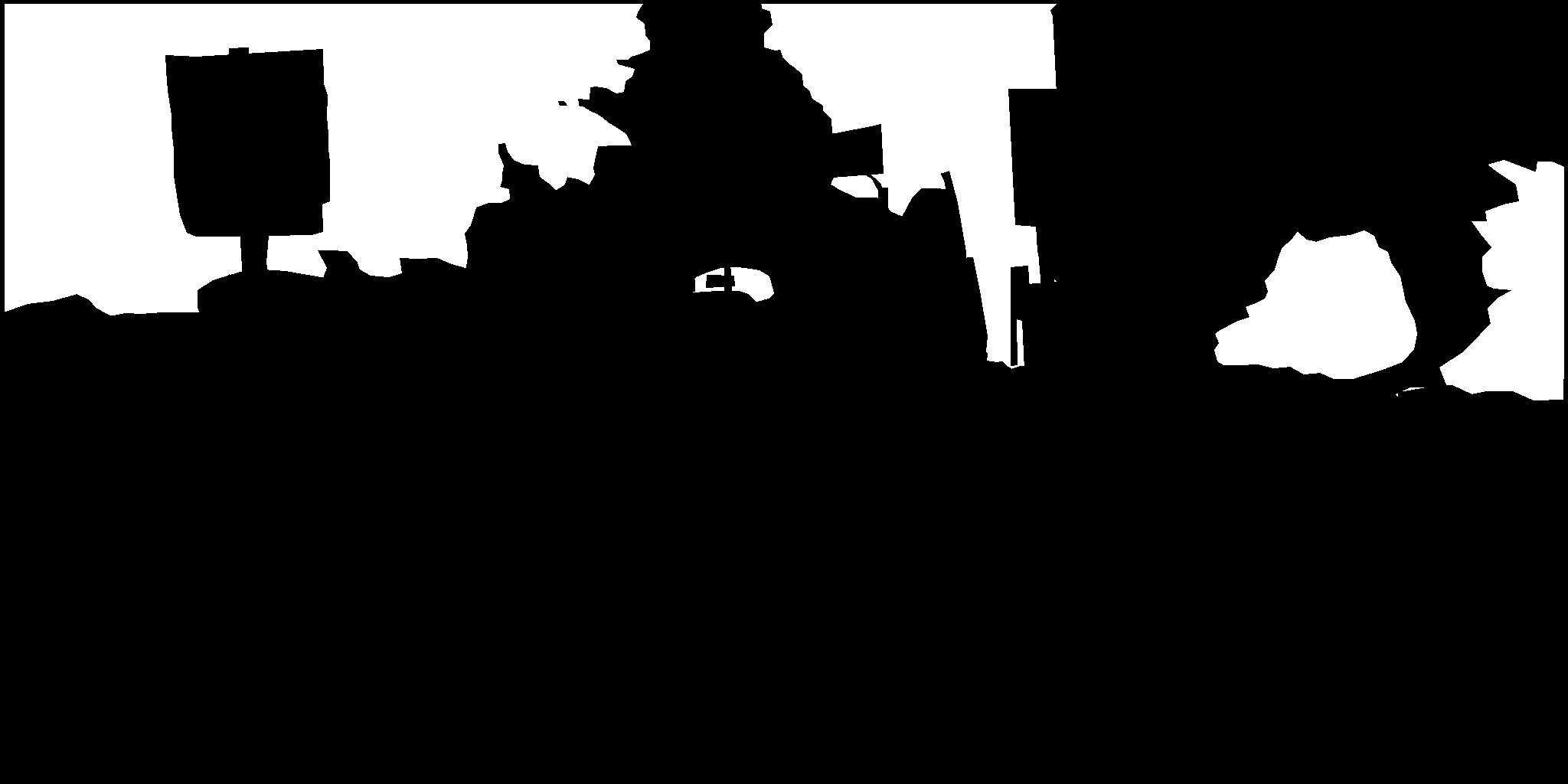}}\\
	\vspace{1mm}
	\resizebox{0.45\columnwidth}{!}{%
		\includegraphics[scale=0.5]{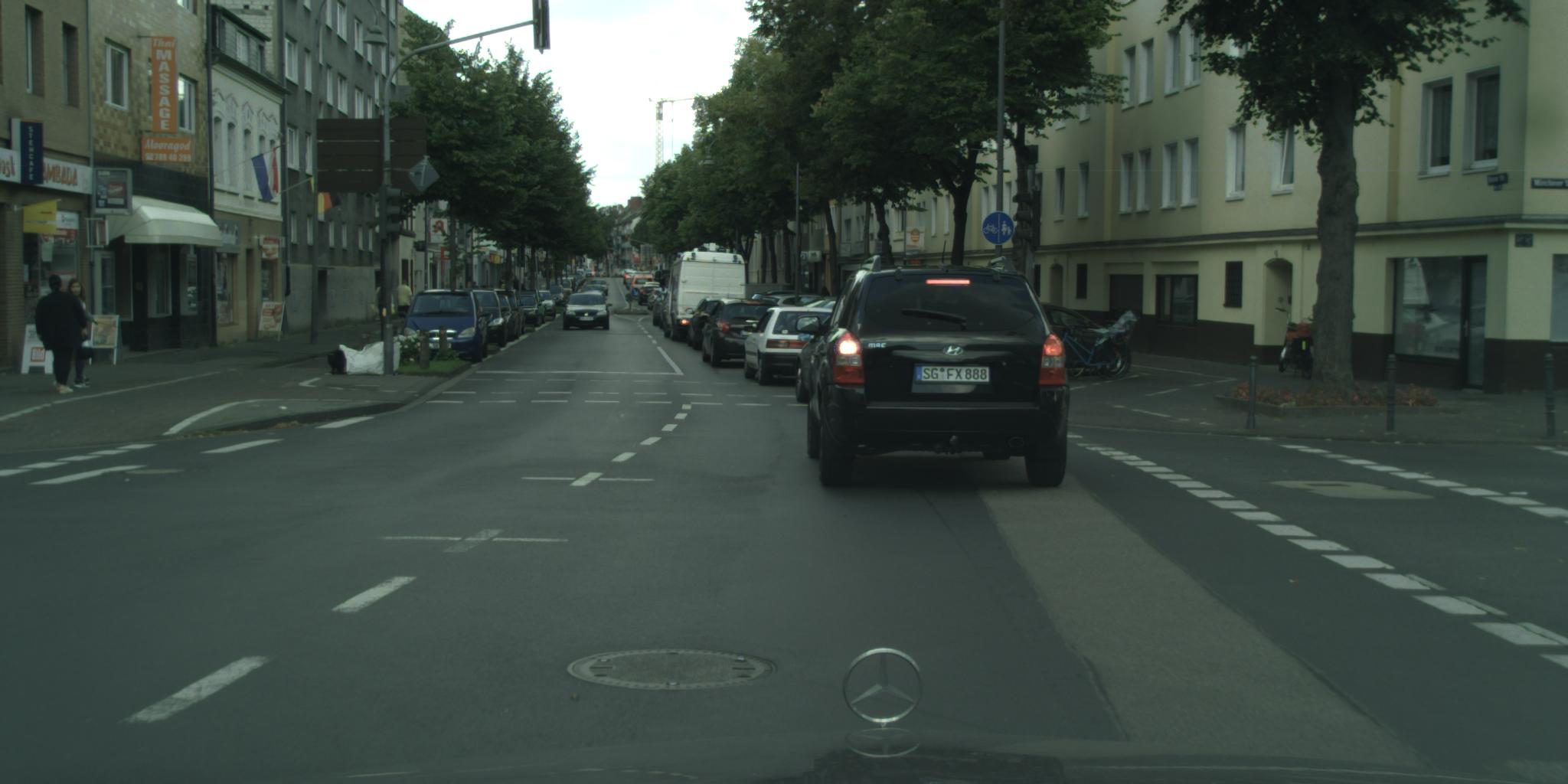}}
	\resizebox{0.45\columnwidth}{!}{%
		\includegraphics[scale=0.5]{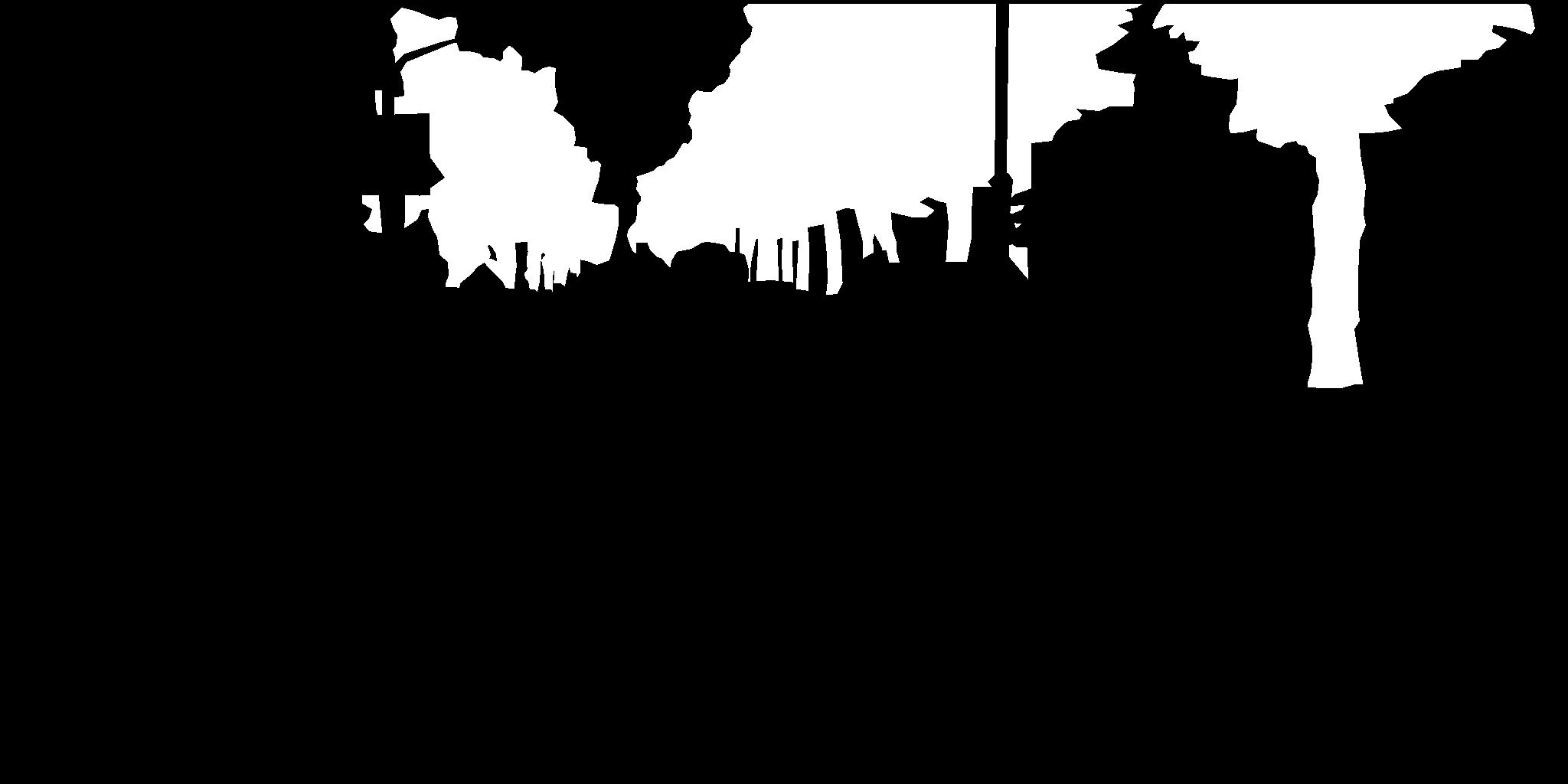}}
	\caption{Two sample images from Cityscapes dataset and their associated vegetation labels}
\end{figure}
\subsection{Evaluation metrics}
We propose two evaluation metrics to compare tree cover estimation: mean IoU for measuring the accuracy of the location of labelled vegetation labels, and Mean Absolute Error for measuring the accuracy in estimating overall GVI. The definitions of the two metrics are provided:
\subsubsection{Mean IoU}
\begin{align*}
n &= \text{number of images in test set}\\
TP_i & = \text{true positive predicted vegetation labels for image i} \\
FP_i & = \text{false positive predicted vegetation labels for image i}\\
FN_i & = \text{false negative predicted vegetation labels for image i}\\
IoU_i &= \frac{TP_i}{TP_i+FP_i+FN_i},\;\;
\bar{IoU} = \frac{1}{n}\cdot\sum_{i=1}^n IoU_i
\end{align*}
\subsubsection{Mean Absolute Error}
\begin{align*}
n &= \text{number of images in test set}\\
M &= \text{number of pixels in a single image}\\
V_i &= \text{binary variable for pixel i, 1 indicates actual vegetation,}\\
&\quad\text{0 indicates actual non-vegetation}\\
\hat{V_i} &= \text{binary variable for pixel i, 1 indicates predicted vegetation,}\\
&\quad\text{0 indicates predicted non-vegetation}\\
\delta_j &= \frac{1}{M}\sum_{i}^M I(\hat{V_i}==1) - I(V_i==1),\;\;
\bar{\delta} = \frac{1}{n}\cdot\sum_{j=1}^n |\delta_j|
\end{align*}
\begin{table*}[h]
	\centering
	\small
	
	\begin{tabular}{|c|ccc|}
		\hline
		Model & Method & Training and Calibration & Prediction Output \\
		\hline
		Benchmark \textit{Treepedia} & Mean shift unsupervised & No training, calibrated by & Fully pixel-segmented \\
		unsupervised segmentation & segmentation \cite{comaniciu2002mean} & Li. et al \cite{li2015assessing} on other GSV images & GSV image \\
		&&&\\
		DCNN semantic & Pyramid scene parsing & Pre-trained on full Cityscapes dataset, &Fully pixel-segmented \\
		segmentation & network (PSPNet) \cite{zhao2017pyramid} & then trained on modified Cityscapes dataset, & GSV image\\
		& with 65,818,363 parameters &and finally on our labelled GSV dataset&\\
		&&&\\
		DCNN end-to-end & Deep residual & Pre-trained on ImageNet dataset, & Single real-valued GVI \\
		& network (ResNet) \cite{he2016deep} & then trained on modified Cityscapes dataset, & number between 0 and 1 \\
		& with 28,138,601 parameters &and finally on our labelled GSV dataset & \\
		\hline
	\end{tabular}
	\caption{Summary of methods, training and calibration techniques and prediction output of models}
\end{table*}
\subsection{Models}
\subsubsection{Benchmark Treepedia unsupervised segmentation}
We use Li. et al's original method as a benchmark metric to compare our performance against. Li. et al uses a mean shift segmentation algorithm that identifies "patches" in images with similar color characteristics \cite{comaniciu2002mean}. Following the same calibrated threshold that the original \textit{Treepedia} project used, the benchmark unsupervised segmentation method identifies "green" patches and labels them as vegetation pixels. Using this method we can see that green but non-tree objects, such as grass or green-colored objects like green shipping containers, can be misclassified (top image of Fig. 5).
\begin{figure}[h]
	\centering
	\resizebox{\columnwidth}{!}{%
		\includegraphics[scale=0.5]{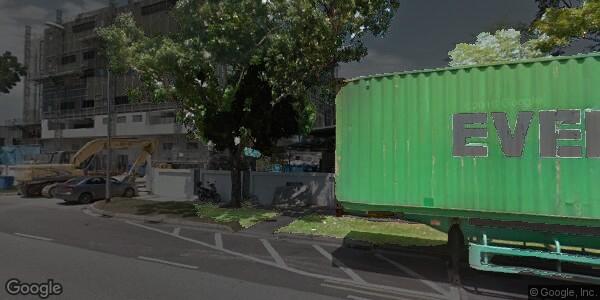}}
	\small{Actual GVI: 21.9\%, Estimated GVI: 40.6\%\\Absolute Error: 18.7\%, IoU: 32.4\%}\\
	\vspace{1mm}
	\resizebox{\columnwidth}{!}{%
		\includegraphics[scale=0.5]{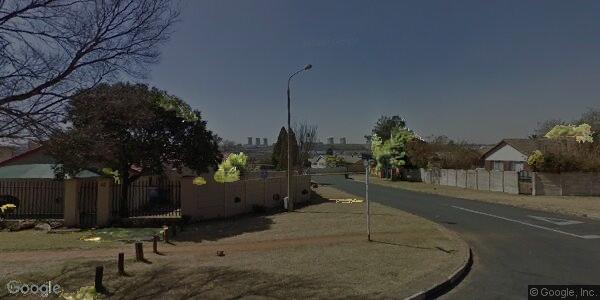}}
	\small{Actual GVI: 25.0\%, Estimated GVI: 3.07\%\\Absolute Error: 21.9\%, IoU: 7.36\%}
	\caption{Example results from segmentation algorithm from benchmark unsupervised segmentation method. Highlighted areas are patches labelled as vegetation with this method.}
\end{figure}
\subsubsection{DCNN semantic segmentation}
Semantic segmentation is a supervised learning problem. Therefore, unlike Li. et al's unsupervised approach, each pixel or patch of pixels is assigned a label of semantic meaning learned by the DCNN from prior training. We adopt Zhao. et al's \cite{zhao2017pyramid} Pyramid Scene Parsing Network (PSPNet)'s architecture for this purpose. PSPNet won the 2016 ImageNet Scene Parsing Challenge and is currently regarded as state-of-the-art for semantic segmentation. We first use pre-trained weights from the original PSPNet trained on the original Cityscapes datasets with its 19 class labels. We then remove the 19 sigmoid units in the last layer and replaced them with 2 sigmoid units. We then pre-train the network again on the aforementioned transformed Cityscapes dataset with only binary labels for vegetation a and non-vegetation. Finally, we train the network on the small labelled GSV dataset itself. Qualitatively, we see that the DCNN semantic segmentation model avoids misclassifying objects as vegetation (top image of Fig. 6).
\begin{figure}[h]
	\centering
	\resizebox{\columnwidth}{!}{%
		\includegraphics[scale=0.5]{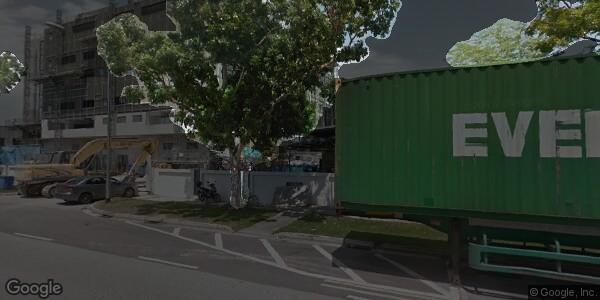}}
	\small{Actual GVI: 21.9\%, Estimated GVI: 22.1\%,\\Absolute Error: 0.20\%, IoU: 89.8\%}\\
	\vspace{1mm}
	\resizebox{\columnwidth}{!}{%
		\includegraphics[scale=0.5]{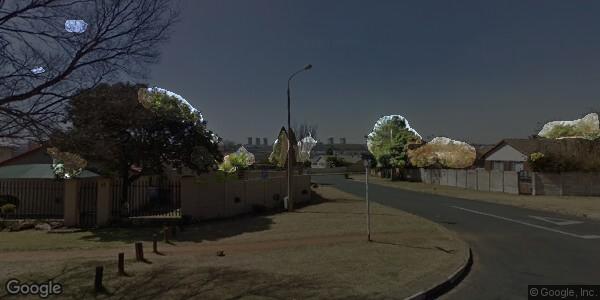}}
	\small{Actual GVI: 25.0\%, Estimated GVI: 5.22\%,\\Absolute Error: 19.78\%, IoU: 20.4\%}
	\caption{Example results from DCNN semantic segmentation algorithm adapted from PSPNet. Highlighted areas are patches labelled as vegetation with this method.}
\end{figure}
\subsubsection{DCNN end-to-end learning} As an alternative approach to the pixel-wise segmentation of vegetation as described above to ultimately compute GVI, we directly estimate GVI with a DCNN model. To conduct end-to-end direct learning of a single GVI value for each image, we adapt He et al's \cite{he2016deep} deep residual network (ResNet) architecture. Since He et al \cite{he2016deep}, other network architectures including PSPNet have widely adopted residual connections to allow for effective feature learning through a high number of convolutional layers. Using a 50 layered ResNet as the base architecture, we add 3 more layers of dense connections at the end, with a final layer consisting of a single sigmoid unit. We first initialize the network with weights that have been pretrained on the ImageNet dataset. We then pre-train the modified ResNet with the transformed Cityscapes dataset and associated true GVI labels, before training on the small labelled GSV dataset. It is worth noting that our DCNN end-to-end model has 28,138,601 parameters, as compared to our DCNN semantic segmentation model, which has 65,818,363 parameters. At testing time, this difference implies that DCNN end-to-end model would outperform DCNN semantic segmentation as the reduced model complexity translates to an increased number of GSV images per evaluation batch.
\subsection{Using Grad-CAM to visualize the DCNN end-to-end model}
The lack of an intermediate image segmentation mask makes it difficult to interpret the learning process of the DCNN end-to-end model. It becomes particularly difficult to confirm or communicate the features that the DCNN end-to-end model has learned in order to estimate GVI. Selvaraju et al \cite{selvaraju2016grad} developed Grad-CAM in order to produce visual explanations for features learned in particular convolutional layers. Through partial linearization and taking the rectified linear function of linear combination of feature maps and neural importance weights, we can visualize areas of the original input image that contribute positively to the prediction of a particular class. Selvaraju et al found that DCNN models may learn biases that may not be completely generalizable and may introduce unwanted prejudice, thus visualizing the active features can help identify those biases. Hence, we apply Grad-CAM to our DCNN end-to-end model to understand whether our model has learned generalizable features.
\begin{figure}[h]
	\centering
	\resizebox{\columnwidth}{!}{%
		\includegraphics[scale=0.5]{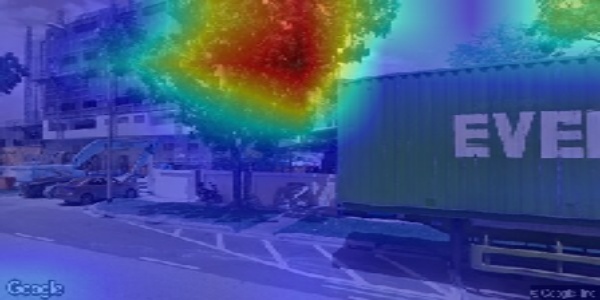}}
	\small{Actual GVI: 21.9\%, Estimated GVI: 27.5\%,\\
		Absolute Error: 5.55\%}\\
	\vspace{1mm}
	\resizebox{\columnwidth}{!}{%
		\includegraphics[scale=0.5]{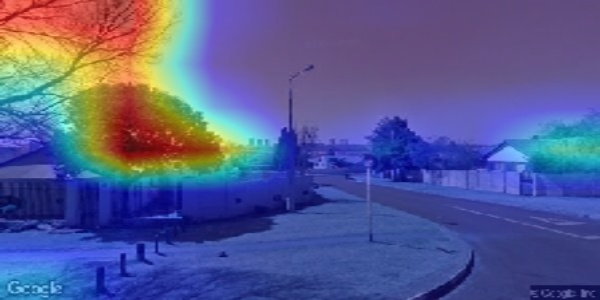}}
	\small{Actual GVI: 25.0\%, Estimated GVI: 21.63\%,\\
		Absolute Error: 3.37\%}
	\caption{Example results from applying Grad-CAM on our trained DCNN end-to-end models. Areas closer to red have a more positive contribution to a higher prediction of GVI than the contribution of areas closer to blue.}
\end{figure}
\begin{table*}[h]
	\centering
	\small
	
	\begin{tabular}{|c|cccc|}
		\hline
		Model & Mean IoU & Mean Absolute & Pearson's Correlation & 5\%-95\% of GVI \\
		& (\%) & Error (\%) & Coefficient & Estimation Error (\%)\\
		\hline
		Benchmark \textit{Treepedia} & 44.7 & 10.1 & 0.708 & -26.6, 18.7\\
		unsupervised segmentation & & &&\\
		DCNN semantic & 61.3 & 7.83 &0.830& -20.0, 12,37\\
		segmentation & & & &\\
		DCNN end-to-end & NA & 4.67 & 0.939&-10.9, 7.97 \\
		\hline
	\end{tabular}
	\caption{Accuracy comparison between models. DCNN end-to-end model does not provide an intermediate image segmentation, \hspace{\textwidth}hence it is not possible to directly evaluate its mean IoU performance.}
\end{table*}
\section{Results}
\subsection{Model Performance}
We find that DCNN models outperform the original \textit{Treepedia} unsupervised segmentation model significantly (Table II). DCNN semantic segmentation achieve a higher Mean IoU and a lower Mean Absolute Error, while DCNN end-to-end achieve the lowest Mean Absolute Error out of all three models. Comparing the model results using the correlation coefficient between model predicted values and the labelled test set values, we find that the DCNN models significantly outperform the original unsupervised segmentation model (Table II). Furthermore, we report the 5\%-95\% of the GVI estimation error to indicate the spread and centrality of the estimation errors. For this metric, we want the spread to be centered on 0 and we want the spread to be small. In this regard, we see that DCNN models outperform the original unsupervised model significantly in prediction accuracy.
\subsection{Interpretability of DCNN end-to-end model}
The Grad-CAM method applied to our trained DCNN end-to-end model and test images suggests that the DCNN end-to-end model learned to identify vertical vegetation in GSV images, without any noticeable systematic biases (Fig. 8). 
\subsection{Running time and scalability}
Training and testing of the DCNN semantic segmentation and DCNN end-to-end learning models were conducted on a system equipped with a single NVIDIA 1080Ti GPU with 11GB of memory. The training of both models took about 48 hours of training time each.

A critical issue that affects the quantification of urban tree cover is the scalability of our GVI algorithm. Cities with complex and large street networks such as London and New York have up to 1 million GSV images. Therefore, fast evaluation time is essential towards the feasibility of large-scale quantification of a city’s tree cover.

As a test of scalability, we tested the speed of our models in processing 100, 1000 and 10000 GSV images, and report the running time in terms of wall-clock time. To speed up computation, we parallelize the Python-based code for the benchmark \textit{Treepedia} unsupervised segmentation, and we increase the evaluation batch size to the maximum that our single GPU system can handle. The ResNet-based DCNN end-to-end model evaluates 10000 GSV images under 40 seconds, which is a 0.1\% of the evaluation time needed for the benchmark \textit{Treepedia} unsupervised segmentation (Table 2). The DCNN end-to-end model is also significantly faster than the DCNN semantic segmentation model, due to the difference in model complexity resulting in a larger evaluation batches for the DCNN end-to-end model. Using linear extrapolation, our fastest model, the DCNN end-to-end model, can calculate the GVI of street-level images in a large city like London with 1 million images in slightly over one hour of processing time.
\begin{table*}[t]
	\centering
	\small
	\begin{tabular}{|c|ccc|}
		\hline
		Model & Running Time for & Running Time for & Running Time for \\
		& 100 images (seconds) & 1000 images (seconds) & 10000 images (seconds) \\
		\hline
		Benchmark \textit{Treepedia} &  39.6 & 381 & 3665 \\
		unsupervised segmentation & & & \\
		DCNN semantic & 20.2 & 207 & 2064 \\
		segmentation & &&\\
		DCNN end-to-end & 1.81& 4.06& 38.9\\
		\hline
	\end{tabular}
	\caption{Running time of models in computing GVI}
\end{table*}
\section{Conclusion}
We demonstrate the potential of deep learning in achieving better performance as measured by GVI correlation, Mean IoU and Mean Absolute Error, as compared to the original unsupervised segmentation models used in the \textit{Treepedia} \cite{treepedia_web} project and website. Notably, our DCNN end-to-end model achieves both significantly better model performance and evaluation speed.

Besides accuracy, we also introduce the use of Grad-CAM to inspect features learned by our DCNN end-to-end model. We find that our trained DCNN end-to-end model accurately learned to pick out vertical vegetation from GSV images without any noticeable biases.

By using street-level GSV images, we are able to utilize the abundance of open-source manually labelled street-level image datasets to heavily pre-train our DCNN models. After pre-training, we tune our DCNN models to GSV images by training on a small-scale dataset.  This saves significant resources that would normally be associated with building a large custom training dataset. Nonetheless, we produced a carefully curated and labelled training set specific to GSV images and vertical vegetation which will be released and hopefully built upon. It provides future researchers a benchmark validation set that they could use to compare their model performance with our models.

With an accurate, interpretable and efficient GVI estimation technique, city planners can apply our models to understand the spatial distribution of urban greenery, and prioritize public tree maintenance and planting efforts. Our technique can also be confidently used by researchers for a wide variety of research topics that relies on accurate estimations of urban greenery or tree cover including environmental inequality \cite{li2015lives} and urban heat island models \cite{mirzaei2010approaches}.

The collection of street-level images from Google, Baidu and Tencent represent one of the few geographically extensive, and informationally rich datasets available to researchers. By using DCNN models, researchers can utilize vast street-level image datasets to learn and understand complex environmental and socio-economic phenomena, in addition to urban tree cover. From setting metrics to measuring evaluation time, we demonstrate the process of rigorously applying deep learning to street-level images. Especially for researchers from fields including GIS and urban studies who lack computational resources, our paper serves as an example of an effective application of deep learning tools in a computationally efficient manner.

\begin{figure*}[b]
	\centering
	\subfloat{
		\resizebox{\columnwidth}{!}{
			\includegraphics[scale=0.2]{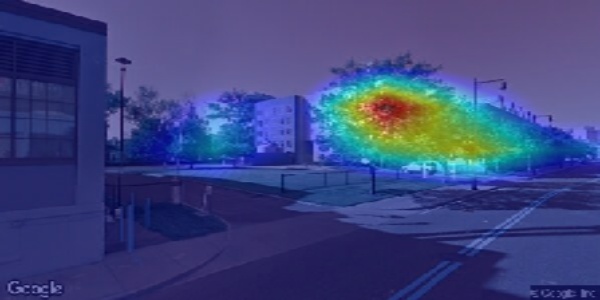}}}
	\subfloat{
		\resizebox{\columnwidth}{!}{
			\includegraphics[scale=0.2]{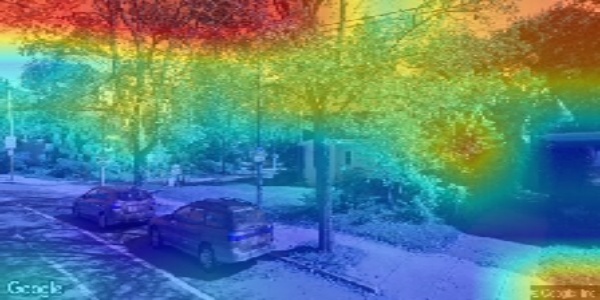}}}\\
	\subfloat{
		\resizebox{\columnwidth}{!}{
			\includegraphics[scale=0.2]{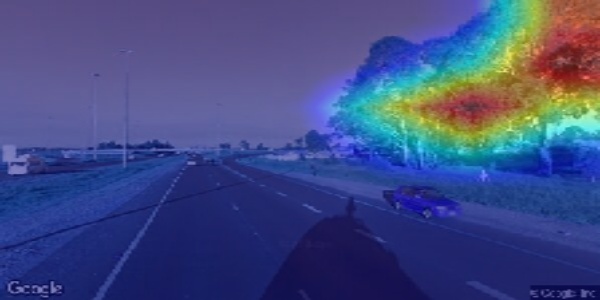}}}
	\subfloat{
		\resizebox{\columnwidth}{!}{
			\includegraphics[scale=0.2]{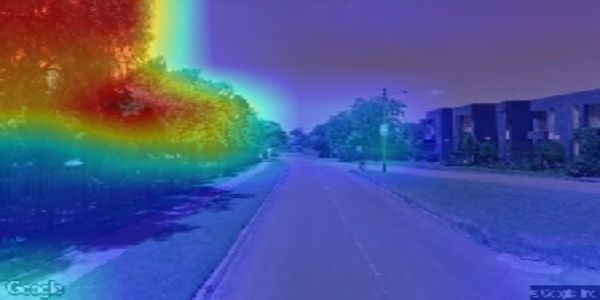}}}\\
	\subfloat{
		\resizebox{\columnwidth}{!}{
			\includegraphics[scale=0.2]{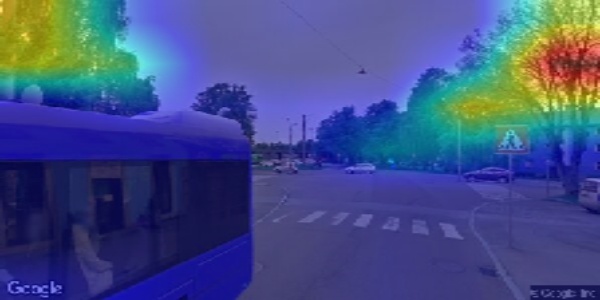}}}
	\subfloat{
		\resizebox{\columnwidth}{!}{
			\includegraphics[scale=0.2]{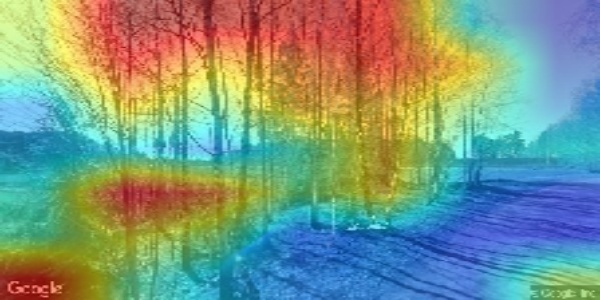}}}\\
	\subfloat{
		\resizebox{\columnwidth}{!}{
			\includegraphics[scale=0.2]{{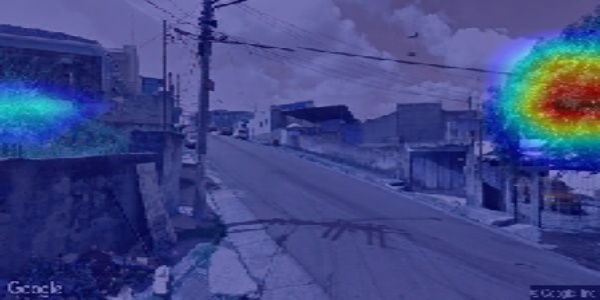}}}}
	\subfloat{
		\resizebox{\columnwidth}{!}{
			\includegraphics[scale=0.2]{{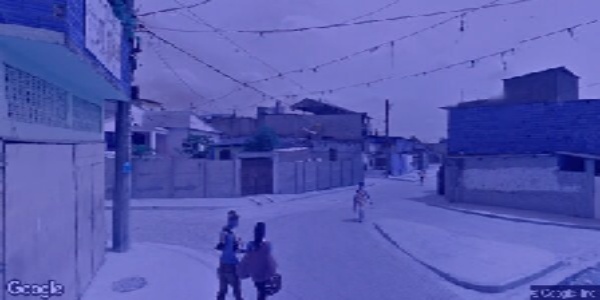}}}}\\
	\subfloat{
		\resizebox{\columnwidth}{!}{
			\includegraphics[scale=0.2]{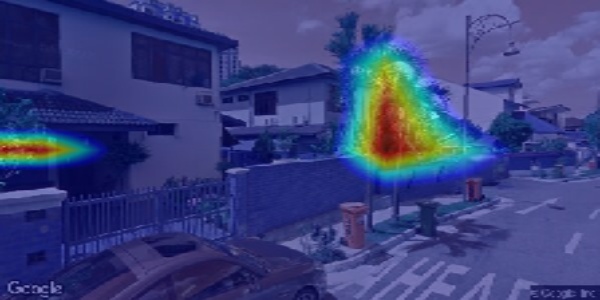}}}
	\subfloat{
		\resizebox{\columnwidth}{!}{
			\includegraphics[scale=0.2]{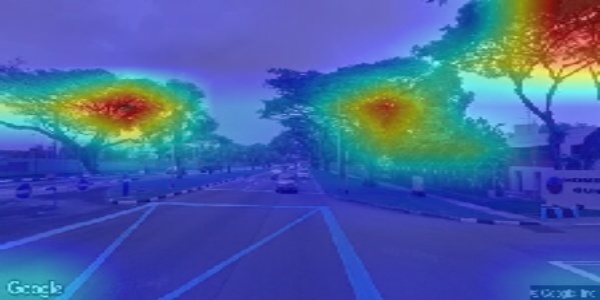}}}\\
	\caption{Results from applying Grad-CAM on our trained DCNN end-to-end modele to understand features learned in the last convolutional layer. 2 images from Cambridge, Johannesburg, Oslo, Sao Paulo, and Singapore are shown in order. Areas closer to red have a more positive contribution to a higher prediction of GVI than the contribution of areas closer to blue.}
\end{figure*}
\section*{Acknowledgment}

The authors would like to thank Fabio Duarte for guidance and support throughout the research process. Bill Yang Cai is supported by a scholarship from the Singapore government. The authors would like to thank Allianz,  Amsterdam Institute for Advanced Metropolitan Solutions, Brose, Cisco, Ericsson, Fraunhofer Institute, Liberty Mutual Institute, Kuwait-MIT Center for Natural Resources and the Environment, Shenzhen, Singapore- MIT Alliance for Research and Technology (SMART), Uber, Victoria State Government, Volkswagen Group America, and all the members of the MIT Senseable City Lab Consortium for supporting this research.




%
\bibliographystyle{IEEEtran}
\bibliography{bib}

\begin{thebibliography}{10}
\providecommand{\url}[1]{#1}
\csname url@samestyle\endcsname
\providecommand{\newblock}{\relax}
\providecommand{\bibinfo}[2]{#2}
\providecommand{\BIBentrySTDinterwordspacing}{\spaceskip=0pt\relax}
\providecommand{\BIBentryALTinterwordstretchfactor}{4}
\providecommand{\BIBentryALTinterwordspacing}{\spaceskip=\fontdimen2\font plus
\BIBentryALTinterwordstretchfactor\fontdimen3\font minus
  \fontdimen4\font\relax}
\providecommand{\BIBforeignlanguage}[2]{{%
\expandafter\ifx\csname l@#1\endcsname\relax
\typeout{** WARNING: IEEEtran.bst: No hyphenation pattern has been}%
\typeout{** loaded for the language `#1'. Using the pattern for}%
\typeout{** the default language instead.}%
\else
\language=\csname l@#1\endcsname
\fi
#2}}
\providecommand{\BIBdecl}{\relax}
\BIBdecl

\bibitem{GUO201627}
Y.~Guo, Y.~Liu, A.~Oerlemans, S.~Lao, S.~Wu, and M.~S. Lew, ``Deep learning for
  visual understanding: A review,'' \emph{Neurocomputing}, vol. 187, pp. 27 --
  48, 2016, recent Developments on Deep Big Vision.

\bibitem{russakovsky2015imagenet}
O.~Russakovsky, J.~Deng, H.~Su, J.~Krause, S.~Satheesh, S.~Ma, Z.~Huang,
  A.~Karpathy, A.~Khosla, M.~Bernstein \emph{et~al.}, ``Imagenet large scale
  visual recognition challenge,'' \emph{International Journal of Computer
  Vision}, vol. 115, no.~3, pp. 211--252, 2015.

\bibitem{Nowak2007220}
D.~Nowak, R.~Hoehn, and D.~Crane, ``Oxygen production by urban trees in the
  united states,'' \emph{Arboriculture and Urban Forestry}, vol.~33, no.~3, pp.
  220--226, 2007.

\bibitem{lafortezza2009benefits}
R.~Lafortezza, G.~Carrus, G.~Sanesi, and C.~Davies, ``Benefits and well-being
  perceived by people visiting green spaces in periods of heat stress,''
  \emph{Urban Forestry \& Urban Greening}, vol.~8, no.~2, pp. 97--108, 2009.

\bibitem{bain2012living}
L.~Bain, B.~Gray, and D.~Rodgers, \emph{Living streets: Strategies for crafting
  public space}.\hskip 1em plus 0.5em minus 0.4em\relax John Wiley \& Sons,
  2012.

\bibitem{jim2006perception}
C.~Jim and W.~Y. Chen, ``Perception and attitude of residents toward urban
  green spaces in guangzhou (china),'' \emph{Environmental management},
  vol.~38, no.~3, pp. 338--349, 2006.

\bibitem{yang2009can}
J.~Yang, L.~Zhao, J.~Mcbride, and P.~Gong, ``Can you see green? assessing the
  visibility of urban forests in cities,'' \emph{Landscape and Urban Planning},
  vol.~91, no.~2, pp. 97--104, 2009.

\bibitem{li2015assessing}
X.~Li, C.~Zhang, W.~Li, R.~Ricard, Q.~Meng, and W.~Zhang, ``Assessing
  street-level urban greenery using google street view and a modified green
  view index,'' \emph{Urban Forestry \& Urban Greening}, vol.~14, no.~3, pp.
  675--685, 2015.

\bibitem{anguelov2010google}
D.~Anguelov, C.~Dulong, D.~Filip, C.~Frueh, S.~Lafon, R.~Lyon, A.~Ogale,
  L.~Vincent, and J.~Weaver, ``Google street view: Capturing the world at
  street level,'' \emph{Computer}, vol.~43, no.~6, pp. 32--38, 2010.

\bibitem{cordts2016cityscapes}
M.~Cordts, M.~Omran, S.~Ramos, T.~Rehfeld, M.~Enzweiler, R.~Benenson,
  U.~Franke, S.~Roth, and B.~Schiele, ``The cityscapes dataset for semantic
  urban scene understanding,'' in \emph{Proceedings of the IEEE conference on
  computer vision and pattern recognition}, 2016, pp. 3213--3223.

\bibitem{comaniciu2002mean}
D.~Comaniciu and P.~Meer, ``Mean shift: A robust approach toward feature space
  analysis,'' \emph{IEEE Transactions on pattern analysis and machine
  intelligence}, vol.~24, no.~5, pp. 603--619, 2002.

\bibitem{seiferling2017green}
I.~Seiferling, N.~Naik, C.~Ratti, and R.~Proulx, ``Green streets- quantifying
  and mapping urban trees with street-level imagery and computer vision,''
  \emph{Landscape and Urban Planning}, vol. 165, pp. 93--101, 2017.

\bibitem{cheng2017remote}
G.~Cheng, J.~Han, and X.~Lu, ``Remote sensing image scene classification:
  benchmark and state of the art,'' \emph{Proceedings of the IEEE}, vol. 105,
  no.~10, pp. 1865--1883, 2017.

\bibitem{wegner2016cataloging}
J.~D. Wegner, S.~Branson, D.~Hall, K.~Schindler, and P.~Perona, ``Cataloging
  public objects using aerial and street-level images-urban trees,'' in
  \emph{Proceedings of the IEEE Conference on Computer Vision and Pattern
  Recognition}, 2016, pp. 6014--6023.

\bibitem{selvaraju2016grad}
R.~R. Selvaraju, M.~Cogswell, A.~Das, R.~Vedantam, D.~Parikh, and D.~Batra,
  ``Grad-cam: Visual explanations from deep networks via gradient-based
  localization,'' \emph{arXiv preprint arXiv:1610.02391}, 2016.

\bibitem{zhao2017pyramid}
H.~Zhao, J.~Shi, X.~Qi, X.~Wang, and J.~Jia, ``Pyramid scene parsing network,''
  in \emph{IEEE Conf. on Computer Vision and Pattern Recognition (CVPR)}, 2017,
  pp. 2881--2890.

\bibitem{he2016deep}
K.~He, X.~Zhang, S.~Ren, and J.~Sun, ``Deep residual learning for image
  recognition,'' in \emph{Proceedings of the IEEE conference on computer vision
  and pattern recognition}, 2016, pp. 770--778.

\bibitem{treepedia_web}
Treepedia: http://senseable.mit.edu/treepedia. Senseable City Lab.

\bibitem{li2015lives}
X.~Li, C.~Zhang, W.~Li, Y.~A. Kuzovkina, and D.~Weiner, ``Who lives in greener
  neighborhoods? the distribution of street greenery and its association with
  residents’ socioeconomic conditions in hartford, connecticut, usa,''
  \emph{Urban Forestry \& Urban Greening}, vol.~14, no.~4, pp. 751--759, 2015.

\bibitem{mirzaei2010approaches}
P.~A. Mirzaei and F.~Haghighat, ``Approaches to study urban heat
  island--abilities and limitations,'' \emph{Building and environment},
  vol.~45, no.~10, pp. 2192--2201, 2010.

\end{thebibliography}

\end{document}